\documentclass[10pt,twocolumn,letterpaper]{article}

\usepackage{wacv}              

\usepackage{graphicx}
\usepackage{amsmath}
\usepackage{amssymb}
\usepackage{booktabs}
\usepackage{multirow,multicol}
\usepackage[accsupp]{axessibility}

\usepackage[pagebackref,breaklinks,colorlinks]{hyperref}

\usepackage[capitalize]{cleveref}
\crefname{section}{Sec.}{Secs.}
\Crefname{section}{Section}{Sections}
\Crefname{table}{Table}{Tables}
\crefname{table}{Tab.}{Tabs.}

\def\wacvPaperID{1222} 
\def\confName{WACV}
\def\confYear{2025}

\begin{document}

\title{Deciphering the complaint aspects: Towards an aspect-based complaint identification model with video complaint dataset in finance}

\author{
    \begin{tabular}[t]{c}
    Sarmistha Das$^{1}$  \quad Basha Mujavarsheik$^{1}$  \quad  
    R E Zera Lyngkhoi$^{1}$ \quad   Sriparna Saha$^{1}$ \\ 
    Alka Maurya$^{2}$ \\
    {$^{1}$Indian Institute of Technology Patna} , {$^{2}$Crisil Limited} \\
    \end{tabular}
}
\maketitle

\begin{abstract}
  In today's competitive marketing landscape, effective complaint management is crucial for customer service and business success. Video complaints, integrating text and image content, offer invaluable insights by addressing customer grievances and delineating product benefits and drawbacks. However, comprehending nuanced complaint aspects within vast daily multimodal financial data remains a formidable challenge. Addressing this gap, we have curated a proprietary multimodal video complaint dataset comprising 433 publicly accessible instances. Each instance is meticulously annotated at the utterance level, encompassing five distinct categories of financial aspects and their associated complaint labels. To support this endeavour, we introduce Solution 3.0, a model designed for multimodal aspect-based complaint identification task. Solution 3.0 is tailored to perform three key tasks: 1) handling multimodal features ( audio and video), 2) facilitating multilabel aspect classification, and 3) conducting multitasking for aspect classifications and complaint identification parallelly. Solution 3.0 utilizes a CLIP-based dual frozen encoder with an integrated image segment encoder for global feature fusion, enhanced by contextual attention (ISEC) to improve accuracy and efficiency. Our proposed framework surpasses current multimodal baselines, exhibiting superior performance across nearly all metrics by opening new ways to strengthen appropriate customer care initiatives and effectively assisting individuals in resolving their problems. 
  
\end{abstract}

\section{Introduction}
In this era of digital transformation, financial complaints have emerged as a significant concern, profoundly affecting both society and organizations. Financial complaint is a formally recorded expression of dissatisfaction concerning fintech products and financial issues, such as transactions, online banking, and loan settlements. Nowadays, regular users are raising their voices through online posted videos as evidence to attract the attention of regulatory organizations overseeing the service and seeking assistance. According to \cite{olshtain198710}, the objective of the complaint is to express the disagreement between both provider and consumer parties. 
Over the past three years, the Indian Finance Ministry has received over 200,000 complaints, as documented in the report available at \footnote{\url{https://sansad.in/getFile/annex/262/AU224.pdf?source=pqars}}. However, as of late 2023, only 2,500 of these complaints had been successfully resolved, highlighting significant inefficiencies and immediate intervention in the resolution process from the organization's end.
In the extensive landscape of financial reviews, discriminating between coherent and rational reviews highlights the disparity between expectation and reality \cite{preotiuc2019automatically}. 
\begin{figure}[]    
\centering
    \includegraphics[scale = 0.38]{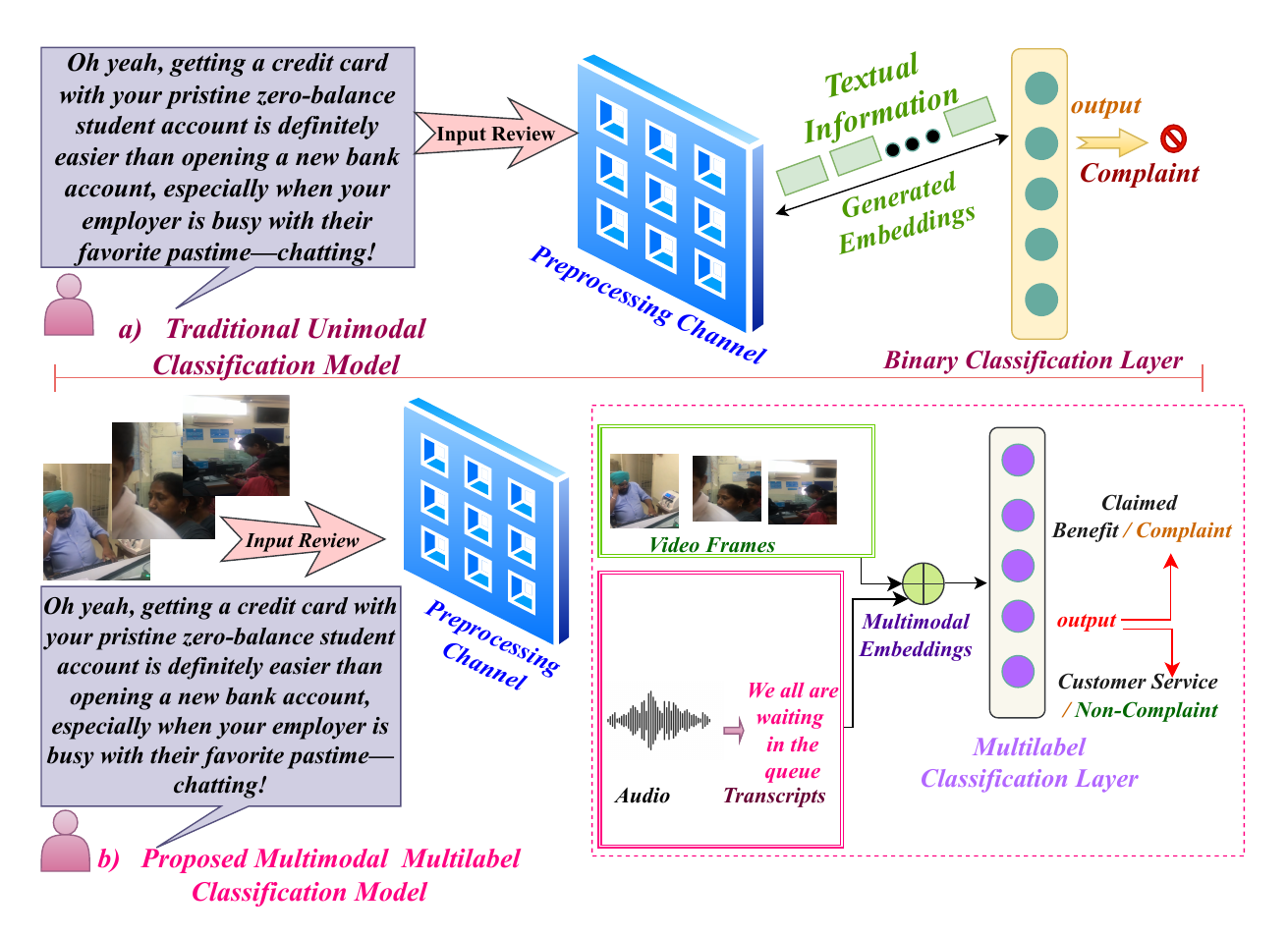}
    \caption{An instance of Aspect Based Complaint identification model vs traditional unimodal model.}\label{situation}    
\end{figure}
Previous research in complaint mining predominantly centered on unimodal review complaints \cite{jin2021modeling}. The true challenge lies in precisely discerning the latent complaint factors that lack explicit elucidation with sufficient explainability. Figure \ref{situation} depicts a situation where a user expresses dissatisfaction with \textbf{\textit{$<XYZ>$ banking services}} without explicitly stating a complaint. Such reviews can confound complaint identification models. Aspect-level complaint identification aims to elucidate the reasons for dissatisfaction, focusing on issues such as \textbf{\textit{inadequate customer service}}. A comprehensive analysis underscores the \textbf{\textit{lack of professionalism on the part of banking employers while customers are standing in a queue}}. The incorporation of multimodal data in financial complaint mining enhances the efficiency of classifier models, enabling responsible organizations to implement more effective customer care initiatives. An effective financial complaint mining model should embody three key qualities. Firstly, it must exhibit robust scalability to handle diverse attributes, including text, images, and lengthy videos, eliminating the need for separate training procedures for each attribute. Secondly, it should excel in inference by effectively extracting implicit evaluations, such as identifying aspects like the lack of professionalism illustrated in Figure \ref{situation}. Lastly, the model should leverage multimodal capabilities to efficiently extract insights from various modalities such as text, audio, and video.

\textbf{\textit{Motivation:}} Past research efforts have leaned on social media analysis to grasp individuals' financial worries \cite{10379488}. However, extant datasets typically tackle this task unimodally, framing it as a classification problem with labels for both complaint and non-complaint instances. To the best of our knowledge, exploring aspect-guided multimodal complaint identification within the financial realm remains unexplored territory. Data resources play a pivotal role as the cornerstone for models of financial complaint mining, where a significant scarcity of multimodal datasets is apparent. Motivated by this identified research gap, we have curated an advanced financial multimodal aspect-oriented complaint dataset \textit{MulComp}. Given that a single video may encompass multiple aspects (as depicted in Figure \ref{situation}), we integrated five prominent financial aspects into our dataset. Consequently, we introduce \textit{Solution 3.0}, a state-of-the-art multimodal aspect-based complaint identification model designed to learn more comprehensive video representations and unseen label representations simultaneously by leveraging a CLIP-based architecture with a distinctive dual frozen encoder.\\
\textbf{\textit{Research Objectives:}} The following constitute the research objectives of the present study:\\
i) Our main objective is to pinpoint the fundamental basis of complaints while investigating the impact of mutiple modalities (text, image and audio) in the realm of financial complaint mining.\\
ii) We seek to evaluate how the robustness of the image segment encoder with contextual attention mechanism can bolster the predictive prowess of our model.\\
iii) Ultimately, we aim to assess the societal benefits stemming from this research endeavour. \\
 \textit{\textbf{Contributions:}} The pivotal contributions of our study are outlined below:\\
i) We introduce the Financial Multimodal Complaint (\textit{MulComp}) dataset, a distinctive dataset encompassing 5 diverse financial aspects, making it unprecedented. Our dataset includes meticulously annotated gold standards labels.
\\ii) Subsequently, we introduce \textit{Solution 3.0}, which addresses three key challenges: a) handling multimodality, b) multilabel aspect classification, and c) multitasking complaint identification with corresponding classified aspects.
\\iii) Additionally, \textit{Solution 3.0} is configured with a novel image segment encoder with contextual attention (ISEC) designed to encode the complex relationships between different labels and dimensions naturally. The proposed model establishes itself as a benchmark for aspect-based complaint identification, surpassing robust and widely used baseline models in both multimodal and unimodal settings in complaint mining tasks.
\section{Related Works}
In 2022, over 14 thousand complaints were made against the companies due to their inferior commitment to resolving performance. Subsequently, SEBI tried to resolve 2838 complaints through platform score in March  2023\cite{Complaints}.
The intricate nature of complaints can lead to ambiguity in representing reviews \cite{jenkins1979levels,radford2019language}. At times, the indication of a complaint is explicit, clearly identifying the party at fault. However, in other instances, the complaint's nature is implicit, lacking explicit blame attribution \cite{reiter2023global}. 

\begin{table}[h]
\centering
\setlength{\tabcolsep}{1pt}
\caption{Comparison of complaint datasets and their associated labels}
\label{datset-comp}
\scalebox{0.57}{

\begin{tabular}{c|c|ccccc|c}
\hline
\multirow{2}{*}{\textbf{Domains}} & \multirow{2}{*}{\textbf{Datasets}} & \multicolumn{5}{c|}{\textbf{Labels}} & \textbf{Multimodality} \\ \cline{3-8} 
  && \multicolumn{1}{c|}{\textbf{Complaint}} & \multicolumn{1}{c|}{\textbf{Severity}} & \multicolumn{1}{c|}{\textbf{Sentiment}} & \multicolumn{1}{c|}{\textbf{Cause}} & \textbf{Aspect} & \textbf{Text/Image/Video} \\ \hline
\multirow{4}{*}{Others}& Complaints \cite{DBLP:conf/acl/Preotiuc-Pietro19a} & \multicolumn{1}{c|}{\checkmark}& \multicolumn{1}{c|}{$\times$}& \multicolumn{1}{c|}{$\times$}& \multicolumn{1}{c|}{$\times$}  & $\times$  & Text  \\ \cline{2-8} 
  & Complaints Severity\cite{jin2021modeling}& \multicolumn{1}{c|}{\checkmark}& \multicolumn{1}{c|}{\checkmark}& \multicolumn{1}{c|}{$\times$}& \multicolumn{1}{c|}{$\times$}  & $\times$  & Text  \\ \cline{2-8} 
  & Complaint ESS \cite{singh2022adversarial}  & \multicolumn{1}{c|}{\checkmark}& \multicolumn{1}{c|}{\checkmark}& \multicolumn{1}{c|}{\checkmark}& \multicolumn{1}{c|}{\checkmark}  & $\times$  & Text  \\ \cline{2-8} 
  & CESAMARD \cite{singh2023knowing} & \multicolumn{1}{c|}{\checkmark}& \multicolumn{1}{c|}{\checkmark}& \multicolumn{1}{c|}{\checkmark}& \multicolumn{1}{c|}{\checkmark}  & \checkmark  & Text+Image\\ \hline
\multirow{2}{*}{Finance}  & X-FinCORP \cite{das2023let}  & \multicolumn{1}{c|}{\checkmark}& \multicolumn{1}{c|}{\checkmark}& \multicolumn{1}{c|}{\checkmark}& \multicolumn{1}{c|}{\checkmark}  & $\times$  & Text  \\ \cline{2-8} 
  & \textit{MulComp}(Proposed)  & \multicolumn{1}{c|}{\checkmark}& \multicolumn{1}{c|}{\checkmark}& \multicolumn{1}{c|}{\checkmark}& \multicolumn{1}{c|}{\checkmark}  & \checkmark  & Video  \\ \hline
\end{tabular}
}
\end{table}

In the field of pragmatics, the work by \cite{olshtain198710} pioneered the classification of complaints into five distinct types: a) Beyond Reproach, b) Explicit Complaint, c) Statement of Disapproval, d) Warning, and e) Allegation. However, Trosborg et al.'s foundational study \cite{trosborg2011interlanguage} identified four primary levels of complaint severity: a) Implicit Reproach Absent, b) Disapproval, c) Accusation, and d) Blame. Gradually, Transformers-based Complaint categorization based on their severity criteria was developed by  \cite{jin2021modeling}. Nonetheless, companies must categorize financial complaints according to their severity and offer comprehensive explanations \cite{10379488}. Subsequently, in 2023, the influence of generative AI was illustrated in the domain of financial unimodal complaint identification, incorporating the provision of causal explanations \cite{das2023let}. A research endeavour was showcased focusing on binary complaint classification in a multimodal information setting, devoid of considering the financial aspect context \cite{singh2022sentiment}. The research presented in \cite{poria2018meld,saha2021towards} has connected vision and language within the associated field of polarity and emotion recognition. The recent multimodal predictive classifier, CMA-CLIP \cite{liu2021cma}, has garnered attention for its ability to predict attributes for predefined classes. However, it faces limitations in handling unseen data. In contrast, the generative model developed by Roy et al.\cite{roy2021attribute} approaches the classification task as a generation task. 
In multilabel classification, the co-occurrence of labels can be effectively modeled using probabilistic graph models \cite{li2016conditional}. However, to address the computational burden of these models, neural network-based solutions have gained prevalence. For instance, Wang et al. \cite{wang2016cnn} used recurrent networks to encode labels into embedding vectors for label correlation modeling, while Lin et al. \cite{lin2018nextvlad} utilized a context-gating strategy to integrate label re-ranking into the network architecture. Additionally, attention mechanisms have been leveraged to model label relationships.\\
\textbf{Our research solution standing:}
Past research explored task-specific layers for complaint detection through fine-tuning pre-trained language models and different attention mechanisms to model multiple-label representations. As of our present understanding, an ultimate comprehensive tool that offers a generalized solution for aspect-based financial multimodal complaints, along with an accompanying dataset, remains absent. We aim to bridge the research gap by presenting our novel model and dataset.\\

\section{Corpus Information}

\begin{figure}[hbt]    
\centering
    \includegraphics[scale = 0.30]{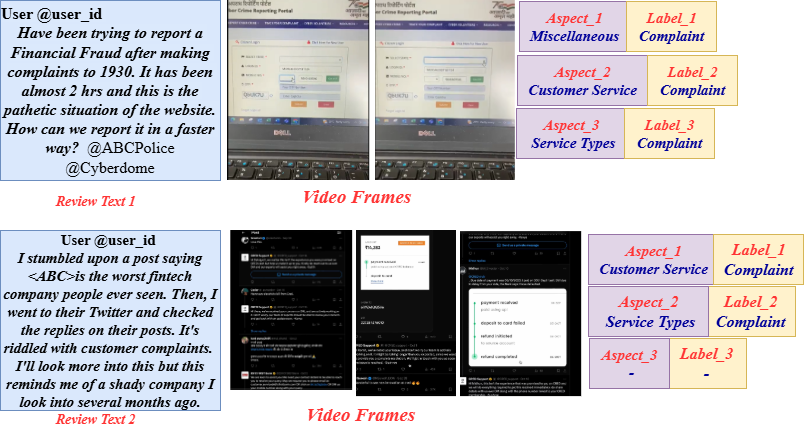}
    \caption{An instance sample in proposed \textit{MulComp} Dataset with two prime class labels}
    \label{example}
\end{figure}


\textbf{Dataset Collection:} During the literature survey, we identified various datasets for complaint detection in general domains, as outlined in Table \ref{datset-comp}. However, in the financial domain, we found only one tailored dataset, X-FinCORP \cite{das2023let}. This dataset is unimodal, which limits its applicability to our specific problem statement. To address this research gap, we curated a pioneering multimodal dataset focusing on aspect-based complaint videos called {\em MulComp}, enriched with gold-standard aspect-based annotations.
\begin{table}[]
\centering
\caption{Finalization of aspects and their overall statistics}\label{tab:datastat}
\scalebox{0.60}{
\begin{tabular}{c|c|c|c}
\hline
\textbf{Final Aspect} & \textbf{Initial Aspects} & \multicolumn{1}{c|}{\textbf{Complaint}} & \multicolumn{1}{c}{\textbf{Non Complaints}}       \\ \hline

\multirow{2}{*}{Customer Service}& Customer Safety          & 17                                      & 41    \\ \cline{2-4}
& Customer Service         & 32                                      & 30  \\ \hline

\multirow{3}{*}{Service Types} & Service Charge           & 59                                      & 134   \\ \cline{2-4}
& Service Types            & 23                                      & 61   \\ \cline{2-4}
& Service Quality          & 24                                      & 35     \\ \hline

Transaction & Transactions             & 29                                      & 75 \\ \hline
Claimed Benefit & Claimed Benefit          & 16                                      & 123                                      \\ \hline
\multirow{3}{*}{Miscellaneous} & Tax Implications         & 10                                      & 12   \\ \cline{2-4}

& Fraud                    & 55                                      & 7 \\ \cline{2-4}
& Others                   & 4                                       & 11      \\ \hline
\end{tabular}}
\end{table}
Initially, we gathered 600 financial video samples from YouTube and Twitter with the explicit goal of creating a publicly accessible resource. To ensure the authenticity of the videos, annotators rigorously verified the background of YouTube channels and their affiliation with financial sectors. Through this process, we filtered for English-language samples, ultimately finalizing 433 complaint review video samples, each validated by a financial industry expert.
\textbf{Dataset Statistics:} Within the {\em MulComp} dataset, each review video averages 5-6 minutes in duration. The identified aspects within these review videos typically span 1-2 minute sections, containing approximately 200 words on average. Out of the 433 videos, 215 are categorized as non-complaints, while 218 are classified as complaints. Table~\ref{tab:datastat} presents detailed statistics concerning the target classes within the {\em MulComp} dataset. To achieve a more balanced representation of our data, we consolidated the 10 original aspects into 5 categories: Miscellaneous, Service Types, Claimed Benefit, Customer Service, and Transaction. 
\textbf{Dataset Annotation:} The manual annotation process for aspect categories and corresponding class labels engaged a team of three annotators. This team comprised a doctoral student from the Computer Science and Engineering department and two undergraduate students, each possessing domain-specific expertise and proficiency in the relevant subject matter. The selection of the undergraduate students underwent a rigorous two-phase competition, open to participants from both the Finance and Computer Science departments. \textbf{Phase-1:} Initially, the expert doctoral student annotated 150 samples, establishing gold standard references. These references were then distributed to all participants, each assigned to annotate an initial set of 50 samples. \textbf{Phase-2:} Subsequently, based on the quality, correctness, and semantic consistency of their annotations, a select group advanced to the last phase, where they annotated an additional 100 samples. Ultimately, only two students emerged as the winners (All the annotators were compensated with per sample 0.5\$ rate).
During the annotation process, selected undergraduate annotators meticulously followed the provided guidelines to ensure comprehensive annotation of the dataset: 1) Annotations were conducted independently of external influence, with a thorough study of referenced samples. 2) Text instances were verified against image instances to identify hidden aspects, if any. 3) Each sample included between one to three aspects, each with a complaint and non-complaint labels. 4) Annotations rectified any biases, ambiguous symbols, or erroneous labels observed.
Ensuring consensus among multiple annotators is crucial when working on annotation tasks involving two or more annotators while creating a reliable annotated dataset. Fleiss' kappa score \cite{fleiss1971measuring}, a standard metric commonly used for this purpose, was employed to evaluate the agreement between the three annotators in our study. The computed Fleiss' kappa scores for the aspect category and complaint label annotation tasks were 0.64, and 0.81, respectively. These scores indicate a high level of agreement among the annotators, suggesting that the annotations are of good quality. We associate only three different aspects for each domain.
\section{Methodology}
This section provides a comprehensive explanation of the problem formulation and outlines the framework. The following sections delve into the key components of the proposed architecture, justifying its adoption.
\begin{figure*}
\centering
\includegraphics[scale=0.36]{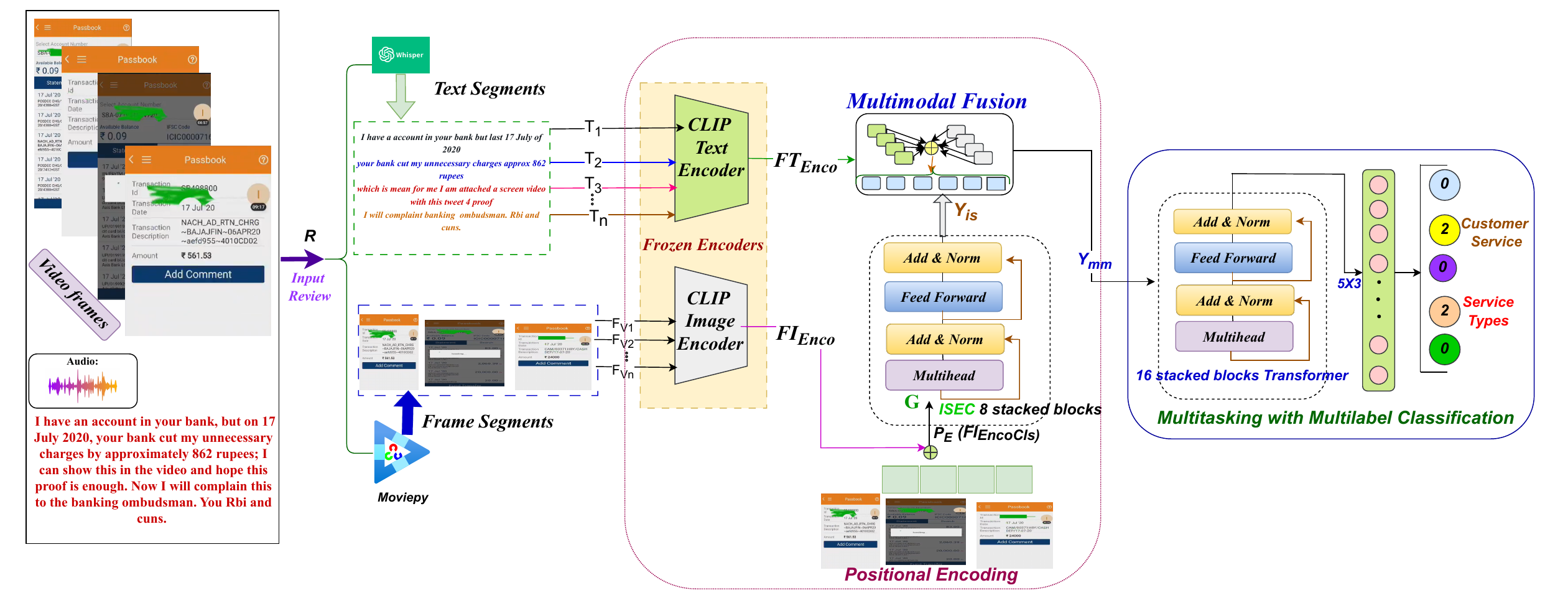}
\caption{ Architectural view of our proposed \textit{Solution 3.0} model; Video frames and Audio Transcripts are passed as input }
\label{fig1}
\end{figure*}
\subsection{Problem Statement}
We conceptualize the financial aspect-based complaint identification model \textit{Solution 3.0} as mentioned in Figure \ref{fig1} as a predictive function that transforms input reviews into derived features, encompassing aspect categories, and target complaint labels.

In this aspect-based complaint identification task, we are provided with a dataset consisting of $n$  reviews, denoted as ${\cal{R}} = \{{\cal{R}}_1, {\cal{R}}_2, \ldots, {\cal{R}}_n\}$. Here, each ${\cal{R}}_i \in {\cal{R}}$ is a review instance that encompasses both acoustic and visual information as $AU_i$, and $V_i$, where $[AU_i, V_i]^n_{i=1}$.

We are also provided with a set of financial aspects denoted as $A = \{a_1, a_2, \ldots, a_m\}$, with $m$ indicating the total number of aspects. Here, $m=5$. The $i^{th}$ financial review ${\cal{R}}i \in {\cal{R}}$ is associated with a set of financial aspects $A^{i} = \{a^{i}_1,a^{i}_2, \ldots, a^{i}_{k_i}\} \subseteq A$, where $k_i$ is the number of aspects associated with ${\cal{R}}_i$. Considering that a single instance may belong to multiple aspect classes, we strategically frame the aspect prediction problem into a multi-label classification task \cite{multilabel_classification}. Therefore, each ${\cal{R}}_i$ sample is labeled with a multi-label aspect vector $Y^i \in \{0,1,2\}^{m}$.  

\begin{equation}
\label{eq:1}\small
{Y_j}^i=
\begin{cases}
    2~,  & ~~ \text{if $a_j \in A^{i}$}\ and\ Complaint \\
    1~,  & ~~ \text{if $a_j \in A^{i}$} \ and\ Non\ Complaint \\
    0~,  & ~~ \text{otherwise}
\end{cases}
\end{equation}
Here, $Y_j$ denotes the $j^{th}$ position in the output aspect. 
We strategies our model to maximize the prediction of aspects and complaint class labels as defined below:

\begin{equation}
    \label{eq:1}
   argmax \ \theta\left (\prod_{i=1}^{n} P(\hat{Y}^i,C_i|{\cal{R}}_i,\theta) \right )
\end{equation}
Where $\theta$ is the optimized parameter. 

\subsection{Pre-Processing Part}
To ensure the quality of information extracted from the input videos (\( R \)), we processed acoustic and visual data separately. Initially, we extracted acoustic information by utilizing the Whisper model and converted it into textual information as transcripts (\( T \)). While video frames (\( F_v \)) were extracted using Moviepy library at a rate of 3 frames per second. To handle the multimodal analysis of large volumes of information, we employed a chunking mechanism, segmenting each video into 2-second intervals. For a given input video \( R \), the chunked representation (\( CR \)) is defined as \( CR \in \{T, F_v\} \), with each segment containing 6 frames (\( 6F_v \)).

\subsection{Phase-1 Feature Extraction}
We employed Whisper to generate the transcript \( T \) from video \( R \), aligned with their timestamps \( t \) according to segments (\( s \)). The final representation of information pairs in chunked reviews is defined as \( CR \in \{T^t_s, {F_v^t}_s\} \). Subsequently, \( CR \) is processed through the pre-trained CLIP model with text and image encoders to generate separate text embeddings (\( T_E \)) and image embeddings (\( I_E \)), denoted as \( CR = CLIP_{encoder}(T_E, I_E) \). 

The CLIP model was fine-tuned by freezing its pre-trained weights. The text processor generates \( T_{Enco} \) and the vision-based processor generates \( I_{Enco} \). To maintain dimensional consistency, padding was applied to both \( T_{Enco} \) and \( I_{Enco} \). The final representations, \( FT_{Enco} \) for text embeddings and \( FI_{Enco} \) for image embeddings, have a shape of \( (b \times cs \times d) \), where \( b \) represents a batch size of 8, \( cs \) denotes chunk size, and \( d \) signifies a 512-dimensional space. Notably, the chunk size can vary based on the length of the video input \( R \).
\subsection{Phase-2 Multimodal Fusion}
To manage the intricate multimodal information, we designed a trainable image segment encoder, which captures dependencies from complex features of the modalities. The image embeddings (\( FI_{Enco} \)) are augmented with a \(\texttt{<cls>}\) token, forming \({FI_{Enco}}_{cls}\). For a given input \( X \), the positional encoding (\( P_E \)) is derived as \( P_x = P_E({FI_{Enco}}_{cls}) \).
\\
To get positional information of each token, we add $P_x$ with \({FI_{Enco}}_{cls}\) to get $G$ as input for the upcoming image segment encoder. 

The given input $G$ is passed into the transformer model where $G \in \mathbb{R}^{k \times d}$. Further, the contextual attention is applied as 
$
\text{SA}(Q, K, V) = \text{Softmax}\left(\frac{QK^T}{\sqrt{d_k}}\right) V
$

The resultant self-attention (SA) provides the dot product of the attention module while maintaining gradient stability by considering key (K), query (Q), and value (V). We further calculate the multi-head attention (MHA) score as follows:
$Y_{is} = MHA(Q, K, V) = LN((Q + Concat(h_1,..., h_n) W^O))$ where $h_i = SA(Q, K, V)$.

Where LN denotes layer normalization, and $W^O$ is the weight matrix. The output of the trainable image segment encoder with contextual attention (IESC) is denoted as \( Y_{is} \). For multimodal fusion, we concatenate the text embeddings (\( FT_{Enco} \)) and \( Y_{is} \), resulting in \( Y_{mm} =  FT_{Enco} \bigoplus Y_{is} \).


\subsection{Phase-3: Multilabel Classification}

\begin{figure}[h]
    \centering
    \includegraphics[scale=0.30]{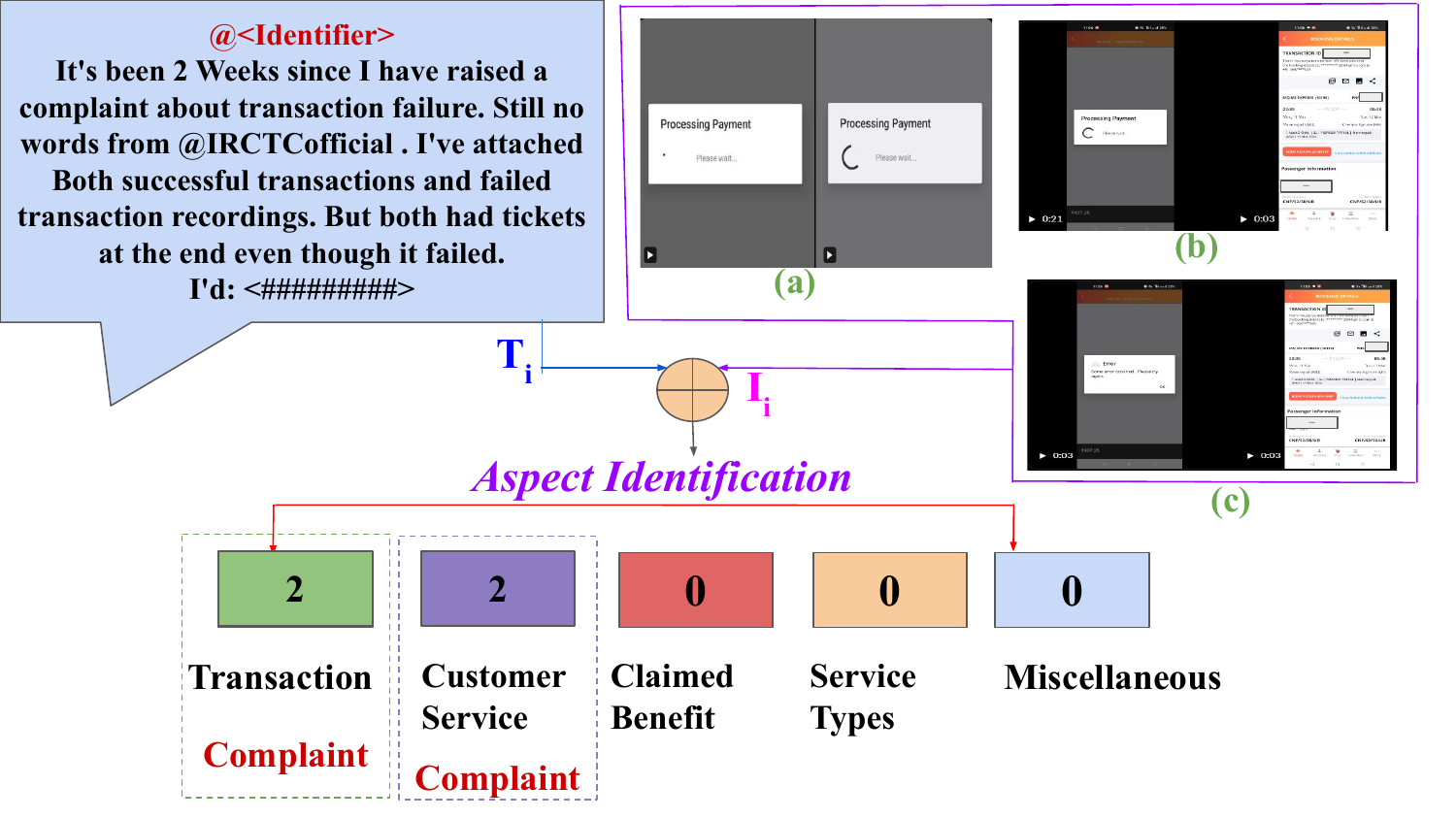}
    \caption{ A sample instance of aspect label classification task}
    \label{multi-example}
    \end{figure}

The fused vector \( Y_{mm} \) is fed into a transformer encoder layer featuring 8 hierarchical attention heads and 16 stacked transformer blocks, followed by a multitask classification layer. This multitask classifier operates on the final video representation, producing an output structured as a \( 5 \times 3 \) matrix. This matrix addresses five classification levels with three possible outcomes for each category: 2 (Aspect present \& complaint), 1 (Aspect present \& non-Complaint) and 0 (No Aspects identified) as mentioned in Figure \ref{multi-example}. The classification process is governed by categorical cross-entropy as the loss function during both training and evaluation.

\section{Experiments, Results, and Analysis}
\textit{This section delves into the experimental setup, provides a concise overview of baselines, presents comparison results, and conducts an analysis of our proposed \textit{Solution 3.0} model.}
We scrutinize the quality of these results through qualitative analysis. Furthermore, we address the shortcomings observed in our model's performance. Our research findings aim to address the following research concerns\footnote{\textbf{Code and data sets are available here: \url{https://github.com/sarmistha-D/Solution-3.O}}}:\\
\textbf{RQ1 :} Can \textit{Solution 3.0} surpass conventional baselines and SOTAs in technical performance and serve as a superior alternative compared to other multimodal fusion methods?\\
\textbf{RQ2:} What interrelationships and effects were noticed among various modalities throughout the experiments? \\
\textbf{RQ3:} How broad is the applicability of \textit{Solution 3.0} in several NLP tasks? What are the societal benefits of \textit{Solution 3.0}?\\
\textbf{RQ4:} What are the limitations of \textit{Solution 3.0}?
\begin{table*}[h]
\centering
\caption{The comparison studies between proposed \textit{Solution 3.0} and other baselines; Here AC and CI stand for Aspect Classification and Complaint Identification task; Statistically significant results are denoted by the symbol "†" where \textit{p} $<$ 0.05 at 5\% significant level.}\label{ablation}
\scalebox{0.57}{
\begin{tabular}{c|c|c|c|c|c|c|c|c|c|c|c|c}
\hline
\multirow{2}{*}{\textbf{Aspects}} & \multirow{2}{*}{\textbf{Task}} & \multirow{2}{*}{\textbf{Metrics}} & \multicolumn{10}{c}{Model Names}         \\ \cline{4-13}

 & &  & \textbf{SOTA\_1} & \textbf{SOTA\_2} & \textbf{VilBERT} & \textbf{LongFormer} & \textbf{Solution 3.0} & \textbf{T5} & \textbf{Gemma} & \textbf{Mistral} & \textbf{GPT3.5} & \textbf{GPT4o} 
 \\ \hline
 
\multirow{4}{*}{Transactions} & \multirow{2}{*}{AC} & Macro-F1 & 50.18 & 
 52.34 & 57.20 & 51.8 & \textbf{57.27}\textsuperscript{†} & 43.12 & 43.50 & 43.44 & 41.43 & 55.58 
 \\ \cline{3-13} 

& & Micro-F1 & 70.19 &  71.58 &  76.46 &  72.19 &  \textbf{76.56}\textsuperscript{†} &  76.81&  76.76 &  76.79 &  42.03 & 57.97\\ \cline{2-13}

& \multirow{2}{*}{CI} & Macro-F1 &  44.22 &  46.17 &  47.54 &  47.92  &  \textbf{48.38}\textsuperscript{†} &  48.11 &  48.11  &  48.52 &  50.71 & 55.19\\ \cline{3-13}

&& Micro-F1 &  89.17 &  91.67 &  90.62 &  92.67  &  \textbf{93.75}\textsuperscript{†} &  92.75 &  92.75  &  92.75  &  68.12 & 71.01\\ \hline

\multirow{4}{*}{Customer Service} & \multirow{2}{*}{AC} & Macro-F1 &  45.81 &  45.23 &  52.59 &  45.81  &  \textbf{51.87}\textsuperscript{†} &  44.35 &  44.80 &  44.35  &  54.39 & 54.61\\ \cline{3-13} 

&& Micro-F1 &  78.92 &  80.92 &  75.00  &  81.30 &  \textbf{81.25}\textsuperscript{†} &  79.71 &  81.16  &  79.71  &  60.87 & 73.91\\ \cline{2-13} 

& \multirow{2}{*}{CI} & Macro-F1 &  45.65 &  43.98 &  46.67 &  48.11 &  \textbf{60.01}\textsuperscript{†} &  47.33 &  47.73  &  47.73 &  48.71 & 47.05\\ \cline{3-13} 

&& Micro-F1 &  87.23 &  89.43 &  87.50 &  \textbf{91.34} &  90.72\textsuperscript{†} &  89.86 &  91.30 &  91.30 &  69.57 & 66.67\\ \hline

\multirow{4}{*}{Claimed Benifits} & \multirow{2}{*}{AC} & Macro-F1 &  81.23 &  83.03 &  86.23 &  \textbf{94.05} &  88.85\textsuperscript{†} &  78.10 &  42.02 &  41.53 &  52.01 & 49.10 \\ \cline{3-13} 

&& Micro-F1 &  84.12 &  88.12 &  89.06 &  \textbf{95.26} &  92.19\textsuperscript{†} &  84.06 &  72.46 &  71.01 &  52.17 & 49.28\\ \cline{2-13}

& \multirow{2}{*}{CI} & Macro-F1 &  50.11 &  47.19 &  49.20 &  50.55 &  \textbf{90.01}\textsuperscript{†} &  49.26 &  49.26 &  49.26 &  53.22 & 67.66\\ \cline{3-13} 

&& Micro-F1 &  89.45 &  80.43 &  94.81 &  92.27 &  \textbf{95.87}\textsuperscript{†} &  97.10 &  97.10 &  97.10 &  76.81 & 91.30 \\ \hline

\multirow{4}{*}{Service Types} & \multirow{2}{*}{AC} & Macro-F1 &  59.23 &  60.51 &  70.15 &  69.42 &  \textbf{70.85}\textsuperscript{†} &  69.14 &  62.21 &  64.53 &  46.92 & 64.77\\ \cline{3-13}

&& Micro-F1 &  60.34 &  61.34 &  71.84 &  71.25 &  \textbf{71.87\textsuperscript{†}} &  71.01 &  63.77 &  69.57 &  47.83 & 66.67\\ \cline{2-13}

& \multirow{2}{*}{CI} & Macro-F1 &  47.52 &  49.86 &  52.77 &  47.02 &  \textbf{51.77}\textsuperscript{†} &  43.44 &  43.44 &  48.12 &  41.53 & 52.26\\ \cline{3-13}

&& Micro-F1 &  66.61 &  68.27 &  70.31 &  69.57 &  \textbf{71.43}\textsuperscript{†} &  76.81 &  76.81 &  75.36 &  71.01 & 75.36\\ \hline

\multirow{4}{*}{Miscellaneous} & \multirow{2}{*}{AC} & Macro-F1 &  49.21 &  52.18 &  58.97 &  56.35 &  \textbf{57.14}\textsuperscript{†} &  45.24 &  45.24 &  45.24 &  47.69 & 53.29\\ \cline{3-13}

&& Micro-F1 &  74.12 &  76.02 &  75.00 &  80.44 &  \textbf{81.25}\textsuperscript{†} &  82.61 &  82.61 &  64.53 &  49.28 & 59.42\\ \cline{2-13}

& \multirow{2}{*}{CI} & Macro-F1 &  73.19 &  76.19 &  61.21 &  62.47 &  \textbf{79.89}\textsuperscript{†} &  46.09 &  46.09 &  46.09 &  45.67 & 45.24\\ \cline{3-13} 

&& Micro-F1 &  81.78 &  82.12 & 81.25 &  87.02 &  \textbf{89.37}\textsuperscript{†} &  85.51 &  85.68 &  82.51 &  84.06 & 82.69\\ \hline
\end{tabular}}
\end{table*}

\begin{table*}[]
\centering
\caption{Modality wise comparison studies between proposed \textit{Solution 3.0} and other unimodal baselines; Here AC and CI stand for Aspect Classification and Complaint Identification task; Highest values for CI and AC tasks are highlighted in bold}\label{CE_task}
\scalebox{0.56}{
\begin{tabular}{l|l|l|rrrrr}
\hline
\textbf{Aspects}                  & \textbf{Task}       & \textbf{Metrics} & \multicolumn{5}{c}{\textbf{Different Combinations of Solution 3.0}}                                                                                                                                                    \\ \hline
\textbf{}                         & \textbf{}           & \textbf{}         & \multicolumn{1}{l|}{\textbf{Video Only}} & \multicolumn{1}{l|}{\textbf{Audio Only}} & \multicolumn{1}{l|}{\textbf{Multimodal}} & \multicolumn{1}{l|}{\textbf{CLIP Frozen}} & \multicolumn{1}{l}{\textbf{Without IESC}} \\ \hline
\multirow{6}{*}{Transactions}     & \multirow{3}{*}{AC} & Macro-F1          & \multicolumn{1}{r|}{46.66}               & \multicolumn{1}{r|}{55.53}               & \multicolumn{1}{r|}{\textbf{57.27}}      & \multicolumn{1}{r|}{43.86}                & 49.49                                      \\ \cline{3-8} 
                                  &                     & Micro-F1          & \multicolumn{1}{r|}{71.88}               & \multicolumn{1}{r|}{74.13}               & \multicolumn{1}{r|}{\textbf{76.56}}      & \multicolumn{1}{r|}{70.13}                & 74.69                                      \\ \cline{3-8} 
                                  &                     & Hamming Loss      & \multicolumn{1}{r|}{0.281}               & \multicolumn{1}{r|}{0.218}               & \multicolumn{1}{r|}{0.234}               & \multicolumn{1}{r|}{0.219}                & 0.203                             \\ \cline{2-8} 
                                  & \multirow{3}{*}{CI} & Macro-F1          & \multicolumn{1}{r|}{47.96}               & \multicolumn{1}{r|}{47.97}               & \multicolumn{1}{r|}{\textbf{48.38}}      & \multicolumn{1}{r|}{48.3}                 & 48.29                                      \\ \cline{3-8} 
                                  &                     & Micro-F1          & \multicolumn{1}{r|}{92.18}               & \multicolumn{1}{r|}{92.19}               & \multicolumn{1}{r|}{\textbf{93.75}}      & \multicolumn{1}{r|}{91.75}                & 93.05                                      \\ \cline{3-8} 
                                  &                     & Hamming Loss      & \multicolumn{1}{r|}{0.078}               & \multicolumn{1}{r|}{0.078}               & \multicolumn{1}{r|}{0.061}               & \multicolumn{1}{r|}{0.066}                & 0.062                                      \\ \hline
\multirow{6}{*}{Customer Service} & \multirow{3}{*}{AC} & Macro-F1          & \multicolumn{1}{r|}{50.91}               & \multicolumn{1}{r|}{\textbf{58.52}}      & \multicolumn{1}{r|}{51.88}               & \multicolumn{1}{r|}{44.83}                & 43.86                                      \\ \cline{3-8} 
                                  &                     & Micro-F1          & \multicolumn{1}{r|}{79.68}               & \multicolumn{1}{r|}{78.13}               & \multicolumn{1}{r|}{\textbf{81.25}}      & \multicolumn{1}{r|}{81.05}                & 78.13                                      \\ \cline{3-8} 
                                  &                     & Hamming Loss      & \multicolumn{1}{r|}{0.203}               & \multicolumn{1}{r|}{0.2188}              & \multicolumn{1}{r|}{0.187}               & \multicolumn{1}{r|}{0.1875}               & 0.218                                      \\ \cline{2-8} 
                                  & \multirow{3}{*}{CI} & Macro-F1          & \multicolumn{1}{r|}{60.01}               & \multicolumn{1}{r|}{60.21}               & \multicolumn{1}{r|}{\textbf{60.02}}      & \multicolumn{1}{r|}{47.54}                & 46.67                                      \\ \cline{3-8} 
                                  &                     & Micro-F1          & \multicolumn{1}{r|}{90.62}               & \multicolumn{1}{r|}{90.63}               & \multicolumn{1}{r|}{\textbf{90.73}}      & \multicolumn{1}{r|}{90.63}                & 87.51                                      \\ \cline{3-8} 
                                  &                     & Hamming Loss      & \multicolumn{1}{r|}{0.093}               & \multicolumn{1}{r|}{0.093}               & \multicolumn{1}{r|}{0.093}               & \multicolumn{1}{r|}{0.0938}               & 0.125                                      \\ \hline
\multirow{6}{*}{Claimed Benefit}  & \multirow{3}{*}{AC} & Macro-F1          & \multicolumn{1}{r|}{83.39}               & \multicolumn{1}{r|}{87.87}               & \multicolumn{1}{r|}{\textbf{88.85}}      & \multicolumn{1}{r|}{42.86}                & 82.79                                      \\ \cline{3-8} 
                                  &                     & Micro-F1          & \multicolumn{1}{r|}{87.5}                & \multicolumn{1}{r|}{91.44}               & \multicolumn{1}{r|}{\textbf{92.29}}      & \multicolumn{1}{r|}{75.02}                & 92.19                                      \\ \cline{3-8} 
                                  &                     & Hamming Loss      & \multicolumn{1}{r|}{0.125}               & \multicolumn{1}{r|}{0.015}               & \multicolumn{1}{r|}{0.078}               & \multicolumn{1}{r|}{0.25}                 & 0.0781                                     \\ \cline{2-8} 
                                  & \multirow{3}{*}{CI} & Macro-F1          & \multicolumn{1}{r|}{82.93}               & \multicolumn{1}{r|}{49.61}               & \multicolumn{1}{r|}{\textbf{90.62}}      & \multicolumn{1}{r|}{49.61}                & 49.61                                      \\ \cline{3-8} 
                                  &                     & Micro-F1          & \multicolumn{1}{r|}{95.17}               & \multicolumn{1}{r|}{\textbf{95.94}}      & \multicolumn{1}{r|}{95.85}               & \multicolumn{1}{r|}{92.44}                & 93.44                                      \\ \cline{3-8} 
                                  &                     & Hamming Loss      & \multicolumn{1}{r|}{0.015}               & \multicolumn{1}{r|}{0.015}               & \multicolumn{1}{r|}{0.0121}              & \multicolumn{1}{r|}{0.015}                & 0.015                                      \\ \hline
\multirow{6}{*}{Service Type}     & \multirow{3}{*}{AC} & Macro-F1          & \multicolumn{1}{r|}{62.67}               & \multicolumn{1}{r|}{70.46}               & \multicolumn{1}{r|}{\textbf{70.85}}      & \multicolumn{1}{r|}{39.62}                & 57.97                                      \\ \cline{3-8} 
                                  &                     & Micro-F1          & \multicolumn{1}{r|}{65.62}               & \multicolumn{1}{r|}{71.78}               & \multicolumn{1}{r|}{\textbf{71.88}}      & \multicolumn{1}{r|}{65.63}                & 60.94                                      \\ \cline{3-8} 
                                  &                     & Hamming Loss      & \multicolumn{1}{r|}{0.343}               & \multicolumn{1}{r|}{0.283}               & \multicolumn{1}{r|}{0.281}               & \multicolumn{1}{r|}{0.348}                & 0.391                                      \\ \cline{2-8} 
                                  & \multirow{3}{*}{CI} & Macro-F1          & \multicolumn{1}{r|}{46.91}               & \multicolumn{1}{r|}{47.46}               & \multicolumn{1}{r|}{\textbf{51.58}}      & \multicolumn{1}{r|}{43.86}                & 48.73                                      \\ \cline{3-8} 
                                  &                     & Micro-F1          & \multicolumn{1}{r|}{65.62}               & \multicolumn{1}{r|}{\textbf{74.04}}      & \multicolumn{1}{r|}{71.44}               & \multicolumn{1}{r|}{71.13}                & 64.06                                      \\ \cline{3-8} 
                                  &                     & Hamming Loss      & \multicolumn{1}{r|}{0.343}               & \multicolumn{1}{r|}{0.265}               & \multicolumn{1}{r|}{0.265}               & \multicolumn{1}{r|}{0.218}                & 0.359                                      \\ \hline
\multirow{6}{*}{Miscallenous}     & \multirow{3}{*}{AC} & Macro-F1          & \multicolumn{1}{r|}{50.91}               & \multicolumn{1}{r|}{\textbf{67.69}}      & \multicolumn{1}{r|}{57.14}               & \multicolumn{1}{r|}{44.83}                & 69.48                                      \\ \cline{3-8} 
                                  &                     & Micro-F1          & \multicolumn{1}{r|}{79.68}               & \multicolumn{1}{r|}{79.69}               & \multicolumn{1}{r|}{\textbf{81.25}}      & \multicolumn{1}{r|}{81.00}                & 74.94                                      \\ \cline{3-8} 
                                  &                     & Hamming Loss      & \multicolumn{1}{r|}{0.203}               & \multicolumn{1}{r|}{20.31}               & \multicolumn{1}{r|}{0.1875}              & \multicolumn{1}{r|}{0.187}                & 0.14                                       \\ \cline{2-8} 
                                  & \multirow{3}{*}{CI} & Macro-F1          & \multicolumn{1}{r|}{52.90}                & \multicolumn{1}{r|}{67.07}      & \multicolumn{1}{r|}{\textbf{79.91}}               & \multicolumn{1}{r|}{45.76}                & 69.47                                      \\ \cline{3-8} 
                                  &                     & Micro-F1          & \multicolumn{1}{r|}{82.81}               & \multicolumn{1}{r|}{81.25}               & \multicolumn{1}{r|}{\textbf{89.38}}      & \multicolumn{1}{r|}{84.38}                & 85.94                                      \\ \cline{3-8} 
                                  &                     & Hamming Loss      & \multicolumn{1}{r|}{0.171}               & \multicolumn{1}{r|}{0.1875}              & \multicolumn{1}{r|}{0.1563}              & \multicolumn{1}{r|}{0.156}                & 0.14                                       \\ \hline
\end{tabular}}
\end{table*}

\begin{table*}[]
\centering
\caption{Qualitative comparison
between best-performing baselines like ViLBERT, T5, Gemma, SOTA\_2,  human annotators and \textit{Solution 3.0} ; {0: No Aspect present; 1: Aspect present \& Non Complaint; 2: Aspect present \&  Complaint}}
\label{tab:qualititive}
\scalebox{.54}
{
\begin{tabular}{l|lllllll}
\hline
\textbf{Given Review} & \multicolumn{7}{l}{\begin{tabular}[c]{@{}l@{}}When you use your debit card for discount based transaction, every time you reach for it, you're exposing the money in your account. The only person is going to get robbed is you.  When you use \\ your debit card, you can use it for the next 50 years, 20 times a day, you will not raise your credit score by that much. And of course, when you use your debit card, you are liable up to a certain\\  amount and it takes a while in order to get that debit card fixed. So when we do post investigations at breaches, and we say to someone, on your incident, what happened? Well, I was in target, \\ but I used a visa card. So I don't know, nothing. I got a, they canceled my card the next day. Two days later, FedEx sent me a new card, and I was the last I heard about it. What about you? I \\ now use the debit card there that took \$3,000 out of my checking account. It took me two months to get my money back. While they said they were investigating, I had to pay my rent, had kids\\  to wish him. Everything I couldn't pay it because they had my money.\end{tabular}} \\ \hline
\textbf{Model Names}  & \multicolumn{1}{l|}{\textbf{VilBERT}}                                                                                                                   & \multicolumn{1}{l|}{\textbf{T5}}                                                                                           & \multicolumn{1}{l|}{\textbf{Gemma}}                                                                                                                & \multicolumn{1}{l|}{\textbf{SOTA\_2}}                                                                                                                                         & \multicolumn{1}{l|}{\textbf{Solution 3.0}}                                                                                                                                      & \multicolumn{1}{l|}{\textbf{Human Annotator}}                                                                                                                                  & \textbf{Actual Label}                                                     \\ \hline
\textbf{Results}      & \multicolumn{1}{l|}{\begin{tabular}[c]{@{}l@{}}[0, 0, 2, 2, 0]\\ $Aspect_3$: Claimed Benefits\\ $Aspect_4$: Service Types\end{tabular}}                     & \multicolumn{1}{l|}{\begin{tabular}[c]{@{}l@{}}[0, 1, 0, 2, 0]\\ $Aspect_4$: Service Types\end{tabular}}                     & \multicolumn{1}{l|}{\begin{tabular}[c]{@{}l@{}}[2, 0, 0,0 , 1]\\ $Aspect_1$: Transaction\\ $Aspect_5$: Miscellaneous\end{tabular}}                     & \multicolumn{1}{l|}{\begin{tabular}[c]{@{}l@{}}[2, 0, 0,2 , 2]\\ $Aspect_1$: Transaction\\ $Aspect_4$: Service Types\\ $Aspect_5$: Miscellaneous\end{tabular}}                     & \multicolumn{1}{l|}{\begin{tabular}[c]{@{}l@{}}[2, 0, 2, 2, 0]\\ $Aspect_1$: Transaction\\ $Aspect_3$: Claimed Benefits\\ $Aspect_4$: Service Types\end{tabular}}                     & \multicolumn{1}{l|}{\begin{tabular}[c]{@{}l@{}}[2, 0, 2, 2, 0]\\ $Aspect_1$: Transaction\\ $Aspect_3$: Claimed Benefits\\ $Aspect_4$: Service Types\end{tabular}}                    & \begin{tabular}[c]{@{}l@{}}[2, 0, 2, 2, 0]\\ $Aspect_1$: Transaction\\ $Aspect_3$: Claimed Benefits\\ $Aspect_4$: Service Types\end{tabular}                    \\ \hline
\end{tabular}
}
\end{table*}
\par\textbf{Brief Introduction of Baselines and SOTAs: }\\
\textbf{Language Model Baselines:} We examine T5 \cite{raffel2020exploring} Mistral \cite{jiang2023mistral}, GPT-3.5 \cite{brown2020language}, GPT 4o \cite{openai_gpt4} and Gemma \cite{team2024gemma} with fine-tuning to assess their effectiveness for solving AC and CI tasks.\\
\textbf{ViLBERT:} Our methodology integrates Visual-BERT \cite{lu2019vilbert}, a collaborative framework designed to acquire a task-independent visual foundation for coupled visual-linguistic information. ViLBERT employs distinct processing streams for visual and linguistic modalities interconnected through co-attentive transformer layers.\\
\textbf{SOTA\_1:} \cite{10379488}: The SOTA\_1 model, as presented by Das et al., employs a hierarchical attention-based multitasking neural network architecture tailored for detecting financial complaints within a unimodal setting. In our investigation, we repurposed this model for aspect-based complaint identification, employing an identical experimental setup to our current study and leveraging its capacity for multilabel severity classification.\\
\textbf{SOTA\_2} \cite{singh2023knowing}: 
SOTA\_2 is introduced as a multimodal deep learning model employing a bi-transformer fusion technique, aspect-based complaint detection (ABCD), in the e-commerce domain. Initially, the model predicts the specific aspects under scrutiny. Then, in the subsequent stage, it determines whether the predicted aspect is linked to a complaint or not.\\
\textbf{Experimental Setup: }\textit{Solution 3.0} configuration comprised 369 data samples for training the model and 64 data samples to test the model. We utilized an NVIDIA RTX A5000 GPU. The Adam optimizer and categorical cross-entropy loss function were employed for efficient weight updates. Hyperparameters included a learning rate of 1e-5 for the image segment encoder and a learning rate of 1e-6 for the multilabel classifier, a batch size of 8, and a dropout rate of 0.2. The training lasted 100 epochs with a train and test setting. Additionally, we fine-tuned the language models with a similar configuration of 50 epochs using adam with a lr=1e-6 and cross-entropy loss, and the last layer was replaced with a classification layer.
 \textbf{Evaluation Metrics:} For multilabel aspect classification, we used Micro-F1, Macro-F1, and Hamming Loss. Definitions of these metrics can be found in \cite{sun2020lcbm}.
\subsection{Results and Discussion}
This section presents the results and analysis for Aspect Categorization (AC) and Complaint Identification (CI) tasks alongside the previously described baselines and subsequently addresses the research questions.
 \begin{table}[]
\centering
\caption{Generalizibilty comparison study between \textit{Solution 3.0} model on different nlp task-specific models on their datasets }\label{compnlp}
\scalebox{0.64}{
\begin{tabular}{c|c|c|c}

\hline
\textbf{Dataset} & \textbf{Modality} & \textbf{F1 score by Models}& \textbf{F-1\textit{Solution 3.0}}\\
\hline

\textbf{CESAMARD}\cite{singh2023knowing} 
& \textbf{T+I }  & 92.92&92.67\\
\hline
\textbf{MELD \cite{poria2018meld}}& \textbf{T+I }  & 61.60&50.29\\
\hline
\textbf{TCRD \cite{deniz-turkish}}& \textbf{T}  & 87.89&84.08\\
\hline
\textbf{X-FinCORP \cite{10379488}}& \textbf{T}  & 92.58&94.16\\
\hline
\end{tabular}
}
\end{table}

\textbf{Answer to RQ1 \textit{Solution 3.0} as the superior alternative:}
We conducted two ablation studies on our \textit{MulComp} corpus to validate our model configuration choices, as detailed in Table \ref{ablation}. The scalability of our model across all aspects of the Complaint Identification (CI) and Aspect Classification (AC) tasks, except for the $Claimed\ Benefit$ aspect. This discrepancy is likely due to differences in sample distribution and the high contextual dependency of $Claimed\ Benefit$ aspects. While Longformer produced competitive results across various aspects, it outperformed our model in the $Claimed\ Benefit$ aspect. SOTA\_1 showed strong performance in CI tasks but fell short in AC tasks. In contrast, SOTA\_2 excelled in both CI and AC tasks due to its problem-oriented architecture but still did not surpass \textit{Solution 3.0}. VilBERT demonstrated competitive performance but lagged by 3\%. Other language models such as T5, Gemma, Mistral, and GPT-3.5 exhibited similar performance levels, except in the AC tasks for claimed benefits and service types, where Mistral and Gemma significantly underperformed. In contrast, the visual language model GPT-4.0 performed markedly better. However, our model surpassed all these models by a significant margin of 3-4\%. Overall, \textit{Solution 3.0} performs admirably, particularly in aspects dominated by prevalent classes.
\textbf{Answer to RQ2 (Assessment of Importance of Multiple Modalities:} We evaluated the performance of \textit{Solution 3.0} across various modality combinations, as presented in Table \ref{CE_task}. Notably, the audio-only modality outperformed the video-only modality, demonstrating a significant increase of over 4-5\% in AC task accuracy for nearly all aspects. However, for the CI task, the improvement was more modest, averaging around 2-3\%. Remarkably, the audio-only modality achieved the highest macro-F1 scores of 58.82 in customer service and 67.69 in \textit{Miscellaneous} for the AC task. Similarly, in the CI task, it attained the highest macro-F1 score of 67.07 in the \textit{Miscellaneous} aspect. In contrast, the multimodal combination generally secured the highest performance metrics across most aspects, though it did not lead in every instance. Nonetheless, the overall performance trend indicated that AC task scores consistently outperformed CI task scores in the multimodal setting. To validate the innovative contribution of our architecture, we also evaluated the model without the ISEC (image segment encoder with contextual attention) mechanism. The ISEC mechanism excels in handling small embeddings during the dimensional conversion process in fusion, thereby enhancing prediction accuracy. Without the ISEC mechanism, the proposed model experienced a significant performance drop of 4-5\%. Furthermore, we investigated the CLIP frozen setting by removing the ISEC mechanism and the 16-block transformer layer in the multitasking classification section. The CLIP frozen setting yielded the lowest scores across both tasks, underscoring the critical importance of our proposed architectural settings.   
\textbf{Answer to RQ3 (Applicability of \textit{Solution 3.0}):} We evaluated our model's effectiveness on both multimodal and unimodal datasets mentioned in Table \ref{compnlp}. For unimodal datasets, we used X-FINCORP \cite{10379488} (financial complaints), Sem-Eval \cite{mohammad2018semeval} (intensity of emotions and sentiment), and TCRD \cite{deniz-turkish} (Turkish e-commerce reviews). For multimodal datasets, we utilized CESAMARD \cite{singh2023knowing} (e-commerce complaints) and MELD \cite{poria2018meld} (emotion recognition in conversations). Our model, \textit{Solution 3.0}, demonstrated significant performance across all datasets, indicating its generalizability. \textit{Solution 3.0} serves as an invaluable solution for financial organizations, enabling them to pinpoint specific aspects and domains requiring necessary actions, ultimately benefiting both consumers and providers in the long term. For instance, a service provider like <XYZ> can identify the need to enhance net banking functionalities to address payment conversion issues faced by many users. This proactive approach demonstrates to customers that their concerns are valued, thereby fostering trust and confidence in both parties.
\textbf{Answer to RQ4 (what limits the model):}
While our model has demonstrated superior performance compared to baseline models in various uni-modal and multi-modal scenarios, it's crucial to acknowledge its potential limitations. Firstly, our model was trained and evaluated exclusively on English-language tweets, necessitating additional training on different languages and code-mixed settings to broaden its applicability. Secondly, handling sarcasm remains challenging due to the limited availability of curated sarcastic samples. \\
Conclusively, the proposed model, \textit{Solution 3.0}, lives up to its name by offering a singular effective solution for three critical tasks: managing multimodal information, performing multilabel aspect classification, and multitasking complaint identification with aspect classification. This highlights the model's robustness and versatility in financial complaint mining.\quad 
\subsection{ Analytical Discussions} 

\textbf{Qualitative Analysis with Human Expertise:} 
In Table \ref{tab:qualititive}, it's evident that the user explicitly reviewed debit card services, emphasizing the perks of excessive transactions with less security and ease of handling fraudulent incidents. Our model accurately predicted three aspects, whereas VilBERT only identified two, the \textit{Claimed Benefit \& Service Types} aspect but failed to identify the \textit{Transaction} aspect. However, SOTA\_2 correctly identified the \textit{Transaction \& Service Types} aspect, yet it failed to recognize the \textit{Claimed Benefit} aspect and misclassified it as \textit{Miscellaneous}. However, T5 and Gemma exhibit significantly poor results by identifying only one aspect and with one misclassified one. Furthermore, we performed a human evaluation on 60 samples from the test samples, but as mentioned in Table \ref{tab:qualititive}, our proposed \textit{Solution 3.0} outperformed all the other models. Additionally, \textit{Solution 3.0} performed a similar prediction to the human annotator and intensified the performance quality validation.  \\ 
\textbf{Error Analysis:}
We reviewed test set samples to assess \textit{\textit{Solution 3.0}} model's predicted AC and CI labels against actual labels, identifying potential instances of compromised performance and misclassification. i) 
\textbf{Aspect Misinterpretations:} In instances involving the \textit{Miscellaneous} and \textit{Service Types} aspects, the \textit{Solution 3.0} model occasionally misclassifies aspect categories and their associated complaint/non-Complaint labels, particularly in sentences containing multiple aspects. For example, in a sentence with various aspects and a higher number of data samples in the training set, such as: \textit{After buying fruits online with my credit card, I noticed unanticipated additional charges applied, which I was unaware of. It appears my credit card company also want the fruits.} The \textit{Solution 3.0} model erroneously labeled the aspect as '\textit{Miscellaneous}.' However, it should be categorized as '\textit{Service Type,}' since the company's actions violated the transparency of user usage. This misinterpretation likely stems from a paucity of training samples in the service-type aspects, leading to model inaccuracies.
ii) \textbf{Additional Misclassification}: The proposed model also occasionally misclassifies samples, particularly under the \textit{Miscellaneous} aspect category. For instance, tax implications, which fall under \textit{Miscellaneous} aspects, were misclassified by the \textit{Solution 3.0} model as \textit{Service Type}. These misclassifications highlight the model's tendency to inaccurately categorize financial decision-based actions as \textit{Service-Type} aspects, suggesting a need for further refinement in training data and model architecture.
\section{Conclusion}
The financial aspect-based complaint identification (\textit{Solution 3.0}) model establishes a new standard in financial video complaint analysis by integrating aspect-level information in a multimodal setup. The model initiates the process by employing an image segment encoder with contextual attention (ISEC) and maps both acoustic-based text and image label features related to financial aspects. Subsequently, we converted this task into a multilabel with a multitasking classification approach, capable of identifying aspects with corresponding complaint labels in parallel. Additionally, to contribute to the research community, we curated a financial multimodal dataset (\textit{MulComp}) with a balanced proportion of complaint and non-complaint multimodal samples. It underscores that a single review may not always have a clear complaint or non-complaint label. Conversely, only a portion of a review segment may carry more than one complaint aspect, while the rest is non-compliant. This research initiative aims to assist organizations in identifying areas for improvement, resolving customer issues, and restoring their trust.
\textbf{Disclaimer: This article integrates diverse data from various open sources. The company and service names mentioned are for social good research purposes, not for commercial promotion. The authors aim neither to defame nor harm any company's image and refrain from expressing personal preferences.}
\section*{Acknowledgement}
This work is a joint collaboration between the Indian Institute of Technology Patna and CRISIL Data Science Limited. The authors also acknowledge the contributions of dataset annotators Sudip Ghosh and Dishani Samanta for their dedication and assistance.
{\small
\bibliographystyle{ieee_fullname}
\bibliography{main}
}

\end{document}